\documentclass{article}

\usepackage[nonatbib,final]{nips_2016}

\usepackage[utf8]{inputenc} 
\usepackage[T1]{fontenc}    
\usepackage{hyperref}       
\usepackage{url}            
\usepackage{booktabs}       
\usepackage{amsfonts}       
\usepackage{nicefrac}       
\usepackage{microtype}      
\usepackage{graphicx}
\usepackage[backend=bibtex,style=numeric]{biblatex}
\usepackage{amsmath}
\usepackage{gensymb}
\usepackage{color}

\usepackage{abstract}

\title{An unexpected unity among methods for interpreting model predictions\vspace{-0.05in}}

\author{
  Scott M. Lundberg \\
  University of Washington\\
  \texttt{slund1@cs.washington.edu} \\
  \And
  Su-In Lee \\
  University of Washington \\
  \texttt{suinlee@cs.washington.edu} \\
}

\begin{document}
\vspace{-0.25in}
\maketitle
\vspace{-0.25in}
\begin{abstract}
  Understanding why a model made a certain prediction is crucial in many data science fields.
  Interpretable predictions engender appropriate trust and provide insight into how the model may be improved. However, with large modern datasets the best accuracy is often achieved by complex models even experts struggle to interpret, which creates a tension between accuracy and interpretability.  Recently, several methods have been proposed for interpreting predictions from complex models by estimating the importance of input features.
  Here, we present how a model-agnostic additive representation of the importance of input features unifies current methods.
  This representation is optimal, in the sense that it is the only set of additive values that satisfies important properties. We show how we can leverage these properties to create novel visual explanations of model predictions. The thread of unity that this representation weaves through the literature indicates 
  that there are common principles to be learned about the interpretation of model predictions that apply in many scenarios.
\end{abstract}

\section*{Introduction}
\vspace{-0.1in}
A correct interpretation of a prediction model's output is extremely important. This often leads to the use of simple models (e.g., linear models) although they are often less accurate than complex models. The growing availability of big data from complex systems has lead to an increased use of complex models, and so an increased need to improve their interpretability.
%
%
%
Historically, models have been considered interpretable if the behavior of the model as a whole can be summarized succinctly. Linear models, for example, have a single vector of coefficients, which describe the relationships between features and a prediction \textit{across all samples}. Although these relationships are not succinctly summarized in complex models, if we focus on a prediction made on \textit{a particular sample}, we can describe the relationships more easily.
Recent model-agnostic methods leverage this property by summarizing the behaviour of the complex models only with respect to a single prediction \cite{ribeiro2016should, vstrumbelj2014explaining}.

Here, we extend a prediction explanation method based on game theory, specifically on the Shapley value, which describes a way to distribute the total gains to players, assuming they all collaborate \cite{vstrumbelj2014explaining}. We show how this method by Štrumbelj et al. can be extended to unify and justify a wide variety of recent approaches to interpreting model predictions (Figure 1). We term these feature importance values \textit{expectation Shapley (ES) values}; because when the model output is viewed as a conditional expectation (of $y$ given ${\bf x}$), these values are equivalent to the Shapley values, i.e., distribution of credit from coalescent game theory. Intriguingly, ES values connect with and motivate several other current prediction explanation methods:

\begin{figure}
  \centering
  \includegraphics[width=1.0\textwidth]{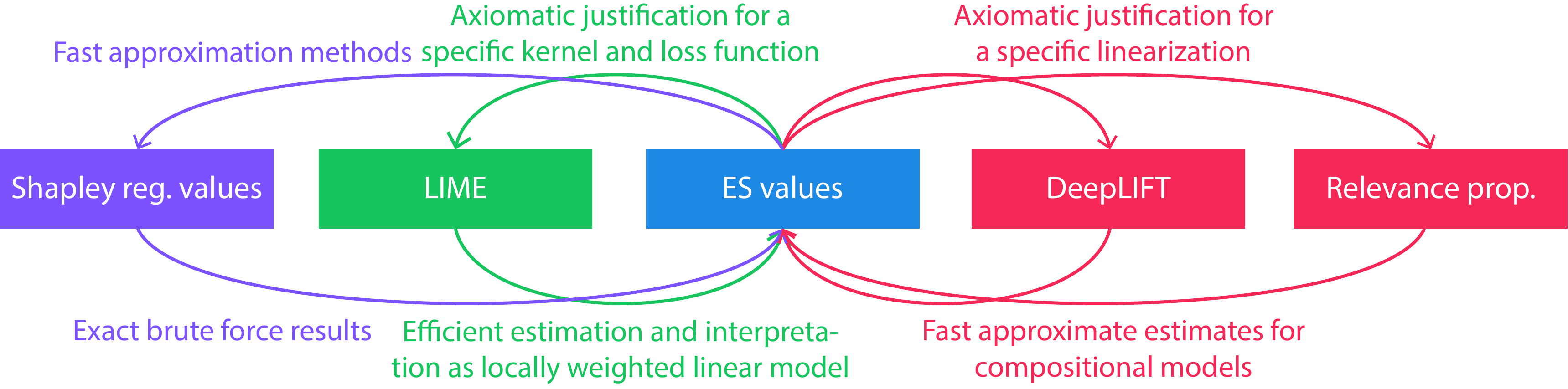}
  \caption{Expectation Shapley (ES) values unify a diverse set of model explanation methods. By connecting these methods with ES values we obtain axiom-based justifications for otherwise arbitrary parameter choices, performance improvements when estimating ES values, and a clearer understanding of how these methods are related.}
\end{figure}

{\bf LIME} is a method for interpreting individual model predictions based on locally approximating the model around a given prediction \cite{ribeiro2016should}. ES values fit into the formalism proposed by LIME and justify a specific local sample weighting kernel. The examples in Ribeiro et al. (2016) \cite{ribeiro2016should} can be viewed as approximations of ES values with a different weighting kernel defining locality.

{\bf DeepLIFT} was recently proposed as a recursive prediction explanation method for deep learning \cite{shrikumar2016not}. DeepLIFT values are ES values for a linearized version of the deep network. This connection motivates the use of DeepLIFT as an extremely efficient sampling-free approximation to ES values. ES values can be also used to uniquely justify specific linearization choices DeepLIFT must make.

{\bf Layer-wise relevance propagation} is another method for interpreting the predictions of compositional models, such as deep learning \cite{bach2015pixel}. As noted by Shrikumar et al., layer-wise relevance propagation is equivalent to DeepLIFT with the reference activations of all neurons fixed to zero \cite{shrikumar2016not}. This implies that layer-wise relevance propagation is also an approximation of ES values, where the primary difference from DeepLIFT is the choice of a reference input to approximate the effect of missing values. By noting that both DeepLIFT and layer-wise relevance propagation are ES value approximations, we can see that DeepLIFT’s proposed improvement over layer-wise relevance propagation is a change that makes DeepLIFT a better approximation of ES values.

{\bf Shapley regression values} are an approach to computing feature importance in the presence of multicollinearity \cite{lipovetsky2001analysis}. They were initially designed to mitigate problems with the interpretability of linear models (although those are typically considered easy to interpret), though they can be applied to other models as well. Shapley regression values require retraining the model on all feature subsets, and can be considered a brute force method of computing ES values. By viewing the model output as an expected value, ES values allow fast approximations in situations where training models on all feature subsets would be intractable.

\section*{Expectation Shapley values and LIME}

Understanding why a model made a prediction requires understanding how a set of interpretable model inputs contributed to the prediction. The original inputs $x \in \mathbb{R}^P$ may be hard for a user to interpret, so a transformation to a new set $x'$ of interpretable inputs is often needed. ES values set $x' = h_x(x)$ to a binary vector of length $M$ representing if an input value (or group of values) is known or missing.
This mapping $h_x$ takes an arbitrary input space and converts it to an interpretable binary vector of feature presence. For example, if the model inputs are word embedding vectors, then $x'$ could be a binary vector of our knowledge of word presence vs. absence. If the model input is a vector of real-valued measurements, $x'$ could be a binary vector representing if a group of measurements was observed or missing.

Prediction interpretation methods seek to explain how the interpretable inputs contributed to the prediction. While the parameters of the original model define this relationship, they do so in a potentially complex manner and do not utilize the interpretable inputs $x'$. To provide interpretability, these methods learn a simple approximation $g(x')$ to the original model for an individual prediction. Inspecting $g(x')$ provides an understanding of the original model’s behavior near the prediction. This approach to local model approximation was formalized recently in Ribeiro et al. as finding an interpretable local model $\xi$ that minimizes the following objective function \cite{ribeiro2016should}:
\begin{equation}
\xi = \mathop{{\arg\min}\vphantom{\sim}}\limits_{\displaystyle _{g \in \mathcal{G}}}  ~ L(f, g, \pi_{x'}) + \Omega(g)
\end{equation}
Faithfulness of the simple model $g(x')$ to the original model $f(x)$ is enforced through the loss $L$ over a set of samples in the interpretable data space $x'$ weighted by $\pi_{x'}$. $\Omega$ penalizes the complexity of $g$.

Given the above formulation for $\xi$ we show the potentially surprising result that if $g$ is assumed to follow the simple additive form:
\vspace{-0.1in}
\begin{equation}
g(x') = \phi_0 + \sum_{i = 1}^M \phi_i x_i',
\end{equation}
where $\phi_i$ (a shortened version of  $\phi_i(f,x)$ when $f$ and $x$ are clear) are parameters to be optimized, then the loss function $L$, the sample weighting kernel $\pi_{x'}$, and the regularization term $\Omega$ are all uniquely determined (up to transformations that do not change $\xi$) given three basic assumptions from game theory. These assumptions are:

\begin{enumerate}
\item {\bf Efficiency}.
\vspace{-0.2in}
\begin{equation}
f(x) = \sum_{i=0}^M \phi_i
\end{equation}
This assumption forces the model to correctly capture the original predicted value.

\item {\bf Symmetry}.  Let $1_S \in \{0,1\}^M$ be an indicator vector equal to $1$ for indexes $i \in S$, and $0$ elsewhere, and let $f_x(S) = f(h_x^{-1}(1_S))$. If for all subsets $S$ that do not contain $i$ or $j$
\begin{equation}
f_x(S \cup \{i\}) = f_x(S \cup \{j\})
\end{equation}
then $\phi_i(f,x) = \phi_j(f,x)$. This states that if two features  contribute equally to the model then their effects must be the same.

\item {\bf Monotonicity}.  For any two models $f$ and $f'$, if for all subsets $S$ that do not contain $i$
\begin{equation}
f_x(S \cup \{i\}) - f_x(S) \ge f'_x(S \cup \{i\}) - f'_x(S)
\end{equation}
then $\phi_i(f,x) \ge \phi_i(f',x)$. This states that if observing a feature increases $f$ more than $f'$ in all situations, then that feature's effect should be larger for $f$ than for $f'$.


\end{enumerate}

Breaking any of these axioms would lead to potentially confusing behavior. In 1985, Peyton Young demonstrated that there is only one set of values that satisfies the above assumptions and they are the Shapley values \cite{young1985monotonic, roth1988shapley}.
ES values are Shapley values of expected value functions, therefore they are the only solution to Equation 1 that conforms to Equation 2 and satisfies the three axioms above. This optimality of ES values holds over a large class of possible models, including the examples used in the LIME paper that originally proposed this formalism \cite{ribeiro2016should}.

\begin{figure}
  \centering
  \includegraphics[width=0.8\textwidth]{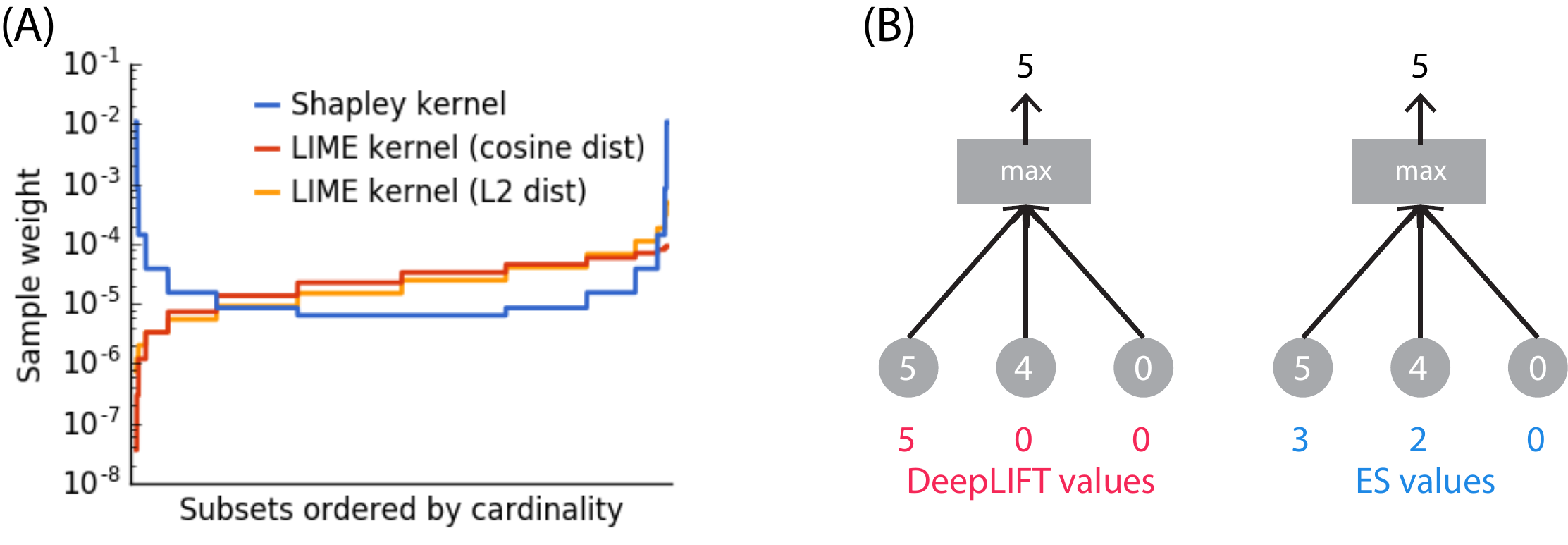}
  \caption{(A) The Shapley kernel for ES values is symmetric, unlike previous heuristically chosen "local" kernels. (B) Rather than making arbitrary linearization choices, ES values justify specific choices for functions. For the max function this choice is different than the one made by DeepLIFT.}
\end{figure}

We found the specific forms of $x'$, $L$, and $\Omega$ that lead to Shapley values as the solution and they are:
\begin{equation}
\begin{split}
\Omega(g) &= 0 \\
\pi_{x'}(z') &= \frac{(M-1)}{(M~choose~|z'|) |z'| (M - |z'|)} \\
L(f,g,\pi_{x'}) &= \sum_{z' \in Z} \left[ f(h_x^{-1}(z')) - g(z') \right]^2 \pi_{x'}(z') \\
\end{split}
\end{equation}
It is important to note that $\pi_{x'}(z') = \infty$ when $|z'| \in \{0,M\}$, which enforces $\phi_0 = f_x(\emptyset)$ and $f(x) = \sum_{i=0}^M \phi_i$. In practice these infinite weights can be avoided during optimization by analytically eliminating two variables using these constraints. Figure 2A compares our Shapley kernel with previous kernels chosen heuristically. The intuitive connection between linear regression and classical Shapley value estimates is that classical Shapley value estimates are computed as the mean of many function outputs. Since the mean is also the best least squares point estimate for a set of data points it is natural to search for a weighting kernel that causes linear least squares regression to recapitulate the Shapley values.

\section*{Expectation Shapley values and DeepLIFT}

DeepLIFT computes the impact of inputs on the outputs of compositional models such as deep neural networks. The impact of an input $x_j$ on a model output $y$ is denoted by $C_{x_j y}$ and
\vspace{-0.07in}
\begin{equation}
\begin{split}
f(x^{(0)}) + \sum_{j=1}^P C_{x_j y} &= f(x)
\end{split}
\end{equation}

\vspace{-0.17in}
where $x^{(0)}$ is a "reference input" designed to represent typical input values. ES value implementations approximate the impact of missing data by taking expectations, so when interpreting $x^{(0)}$ as an estimate of $E[x]$ DeepLIFT is an additive model of the same form as ES values. To enable efficient recursive computation of $C_{x_j y}$ DeepLIFT assumes a linear composition rule that is equivalent to linearizing the non-linear components of the neural network. Their back-propagation rules that define how each component is linearized are intuitive, but arbitrary. If we interpret DeepLIFT as an approximation of ES values, then we can justify a unique set of linearizations for network components based on analytic solutions of the ES values for that component type. One example where this leads to a different, potentially improved, assignment of responsibility is the $max$ function (Figure 2B).

\section*{Visualization of Expectation Shapley values}

Model interpretability is closely tied to human perception. 
We designed a simple visualization based on analogy with physical force (Figure 3A). Each interpretable input $x_i'$ is assigned a bar segment. The width of the segment is equal to the ES value $\phi_i$. Red bar segments correspond to inputs where $\phi_i > 0$, and blue segments to inputs where $\phi_i < 0$. The model output starts at the base value $\phi_0 = f(\emptyset)$ in the center and then is pushed right by the red bars or left by the blue bars in proportion to their length. The final location of the model output is then equal to $f(x) = \sum_{i=0}^M \phi_i$.

\begin{figure}
  \centering
  \includegraphics[width=1.0\textwidth]{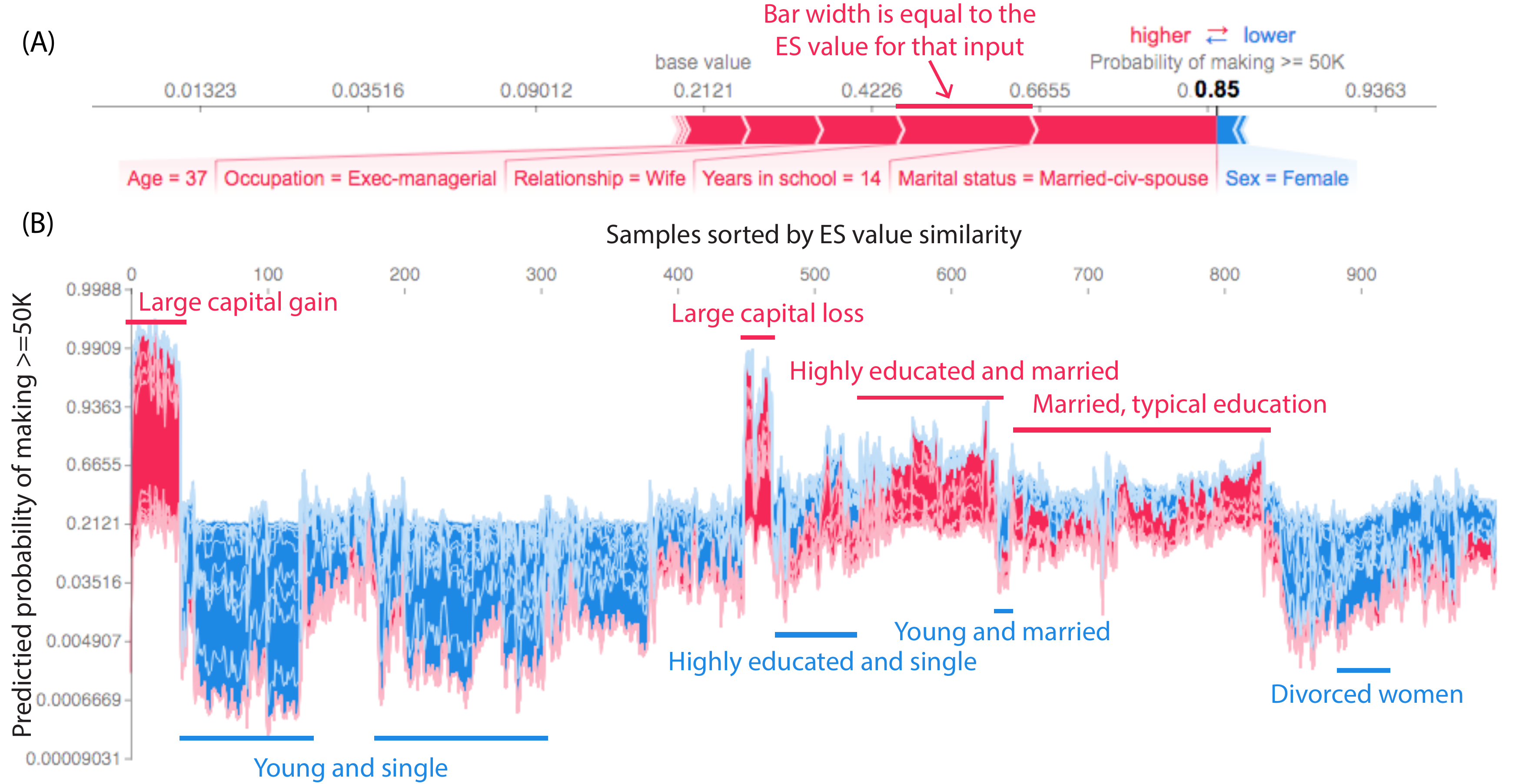}
  \caption{Visual explanations of a model with 2,000 gradient boosted trees on the UCI adult census dataset designed to predict if a person makes >50K annually. (A) A single model output is represented as the sum of input feature ES values. Red positive values push the probability higher to the right, while negative blue values push the probability lower to the left. (B) By turning the individual explanations 90\degree and stacking many together we can see patterns of model behavior in a dataset.}
\end{figure}

While explaining a single prediction is very useful, we often want to understand how a model is performing across a dataset. To enable this we designed a visualization based on rotating the single prediction visualization (Figure 3A) by 90\degree, then stacking many horizontally. By ordering the predictions by explanation similarity we can see interesting patterns (Figure 3B). One such insight for the popular UCI adult census dataset is that marriage status is the most powerful predictor of income, suggesting that many joint incomes were reported, not simply individual incomes as might be at first assumed. For implementation code see \url{https://github.com/slundberg/esvalues}.

\section*{Sample Efficiency and the Importance of the Shapley Kernel}

Connecting Shapley values from game theory with locally weighted linear models brings advantages to both concepts. Shapley values can be estimated more efficiently, and locally weighted linear models gain theoretical justification for their weighting kernel. Here we briefly illustrate both the improvement in efficiency for Shapley values, and the importance of kernel choice for locally weighted linear models (Figure 4).

Shapley values are classically defined by the impact of a feature when it is added to features that came before it in an ordering. The Shapley value for that feature is the average impact over all possible orderings:
\begin{equation}
\phi_i(f,x) = \frac{1}{P!} \sum_{r \in R} \left[ f_x(B_i^r \cup \{i\}) - f_x(B_i^r) \right ]
\end{equation}
where $R$ is the set of all permutations of length $P$, and $B_i^r$ is the set of all features whose index comes before $i$ in permutation $r$. This leads to a natural estimation approach which involves taking the average over a small sample of all orderings \cite{vstrumbelj2014explaining}. While this standard approach is effective in small (or nearly linear) models, penalized regression (using Equation 6) produces much more accurate Shapley value estimates for non-linear models such as a dense decision tree over 10 features (Figure 4A), and a sparse decision tree using only 3 of 100 features (Figure 4B).

While the axioms presented above provide a compelling reason to use the Shapley kernel (Equation 6), it natural to wonder if any reasonable local weighting kernel would produce results similar to the Shapley kernel. It turns out this is not the case, and the Shapley kernel significantly effects how we attribute non-linear effects to various features when compared to the standard exponential kernel used by LIME. For the sparse decision tree used above there is a noticeable change in the magnitude of feature impacts (Figure 4B), and for the dense decision tree we even see the direction of estimated effects reversed (Figure 4A).

\begin{figure}
  \centering
  \includegraphics[width=1.0\textwidth]{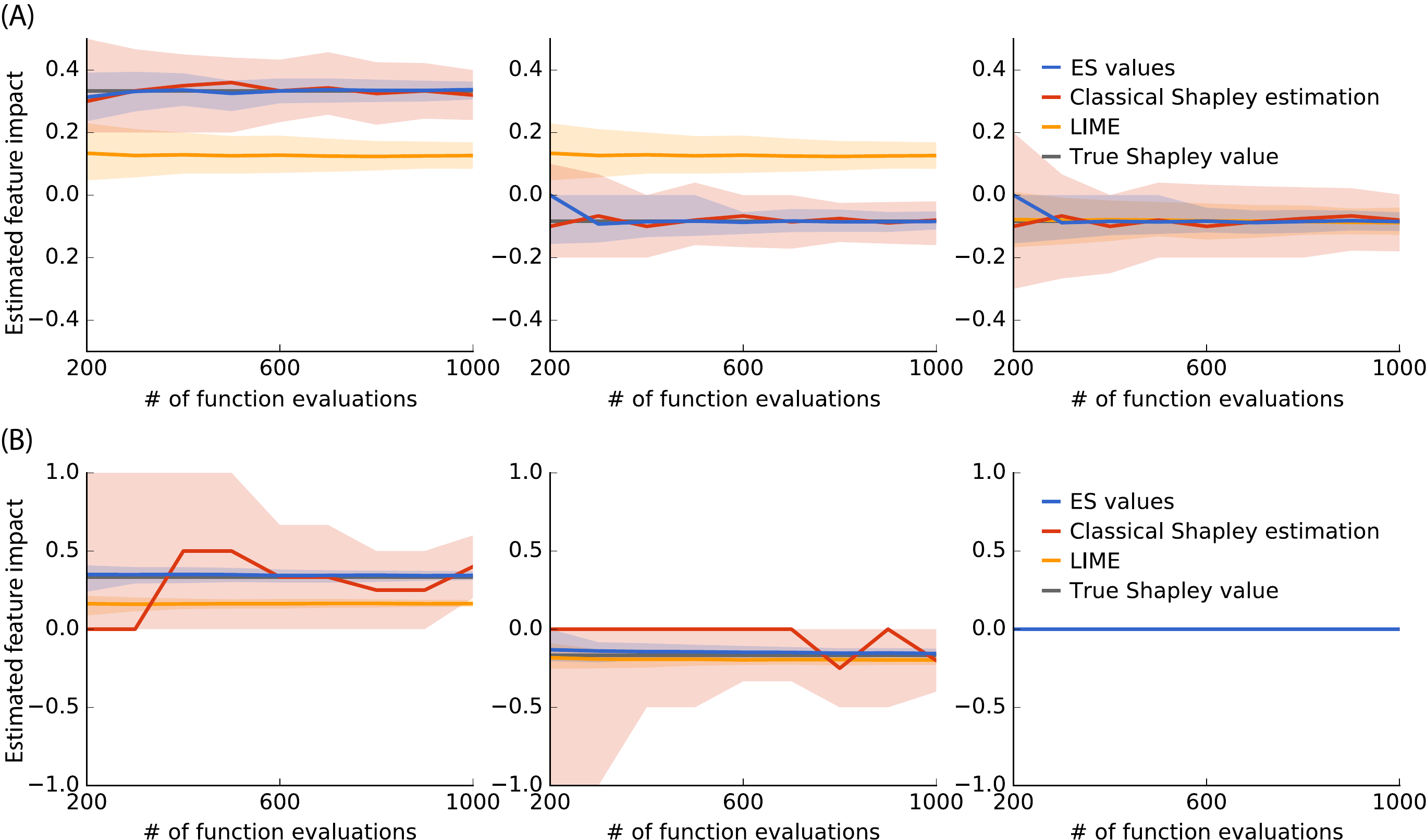}
  \caption{Comparison of three methods for the estimation of feature impacts. ES values are computed used a debiased lasso, classical Shapley estimation uses Equation 8, and LIME uses the open source implementation. Estimates for three different features are shown for each model as the number of evaluations of the original model function is increased. The 10th and 90th percentiles are shown for 200 replicate estimates at each sample size. (A) A decision tree using all 10 input features is explained for a single input.  (B) A decision tree using only 3 of 100 input features is explained for a single input. Two used features and one unused feature are shown.}
\end{figure}
  
\printbibliography

\end{document}